\documentclass[letterpaper, 10 pt, conference]{ieeeconf}  

\IEEEoverridecommandlockouts                              

\usepackage{amsmath}
\usepackage{amssymb}
\usepackage{booktabs}
\usepackage{multirow}
\usepackage{subfigure}
\usepackage{algpseudocode}
\usepackage{amsmath}
\usepackage{algorithm}
\usepackage{float}
\usepackage{microtype}
\usepackage{tabularx}
\usepackage{graphicx}
 \usepackage{hyperref}
\usepackage{cite}
\usepackage{pifont}
\usepackage{float}
\usepackage[export]{adjustbox}
\title{\LARGE \bf
Words to Wheels: Vision-Based Autonomous Driving \linebreak 
Understanding Human Language Instructions Using Foundation Models
}

\author{Chanhoe Ryu$^{1}$, Hyunki Seong$^{1}$, Daegyu Lee$^{2}$, Seongwoo Moon$^{1}$, Sungjae Min$^{1}$ and D.Hyunchul Shim$^{1*}$
\thanks{\textsuperscript{1} Department of Electrical Engineering, Korea Advanced Institute of Science and Technologies (KAIST), Daejeon, Republic of Korea
        {\texttt{\{ryuchanhoe, hynkis, seongwoo.moon, sungjae\_min, hcshim\}@kaist.ac.kr}}}
\thanks{\textsuperscript{2} Department of Mobility Infrastructure, Electronics and Telecommunications Research Institute (ETRI), Daejeon, Republic of Korea
        {\texttt{\{lee.dk\}@etri.re.kr}}}
\thanks{\textsuperscript{*} Corresponding Author}
\thanks{This work was supported by Electronics and Telecommunications Research Institute (ETRI) grant funded by the Korean government (24ZR1210, DNA-Based National Intelligent Core Technology Development).}
}

\begin{document}

\maketitle
\thispagestyle{empty}
\pagestyle{empty}

\begin{abstract}

This paper introduces an innovative application of foundation models, enabling Unmanned Ground Vehicles (UGVs) equipped with an RGB-D camera to navigate to designated destinations based on human language instructions. Unlike learning-based methods, this approach does not require prior training but instead leverages existing foundation models, thus facilitating generalization to novel environments. Upon receiving human language instructions, these are transformed into a `cognitive route description' using a large language model (LLM)—a detailed navigation route expressed in human language. The vehicle then decomposes this description into landmarks and navigation maneuvers. The vehicle also determines elevation costs and identifies navigability levels of different regions through a terrain segmentation model, GANav, trained on open datasets. Semantic elevation costs, which take both elevation and navigability levels into account, are estimated and provided to the Model Predictive Path Integral (MPPI) planner, responsible for local path planning. Concurrently, the vehicle searches for target landmarks using foundation models, including YOLO-World and EfficientViT-SAM. Ultimately, the vehicle executes the navigation commands to reach the designated destination, the final landmark. Our experiments demonstrate that this application successfully guides UGVs to their destinations following human language instructions in novel environments, such as unfamiliar terrain or urban settings.

\end{abstract}

\section{INTRODUCTION}
The primary objective in contemporary robotics is to harness the power of foundation models effectively. These models, invaluable for their ability to generalize from extensive datasets, offer transformative opportunities for enhancing human life through their versatile applications. Foundation models such as large language models (LLMs), vision-language models (VLMs), and image segmentation models are reshaping our understanding of robotic capabilities by enabling robots to process and interpret the world in ways similar to human cognition and perception. This is particularly evident in unmanned ground vehicles (UGVs), which, despite their increasing needs for food delivery, postal services, and surveillance, still face limitations due to their reliance on infrastructure such as GPS and the need for onsite data.

Contrastingly, human navigation often relies solely on verbal instructions and visual cues, using landmarks to guide paths without the need for precise global positioning systems, pre-built maps, or prior knowledge. This insight leads to a crucial research question: ``Can unmanned ground vehicles interpret human language instructions and navigate effectively without conventional infrastructure or pre-acquired knowledge by utilizing foundation models?"
\begin{figure}[t]
  \centering
   \includegraphics[width=0.99\linewidth]{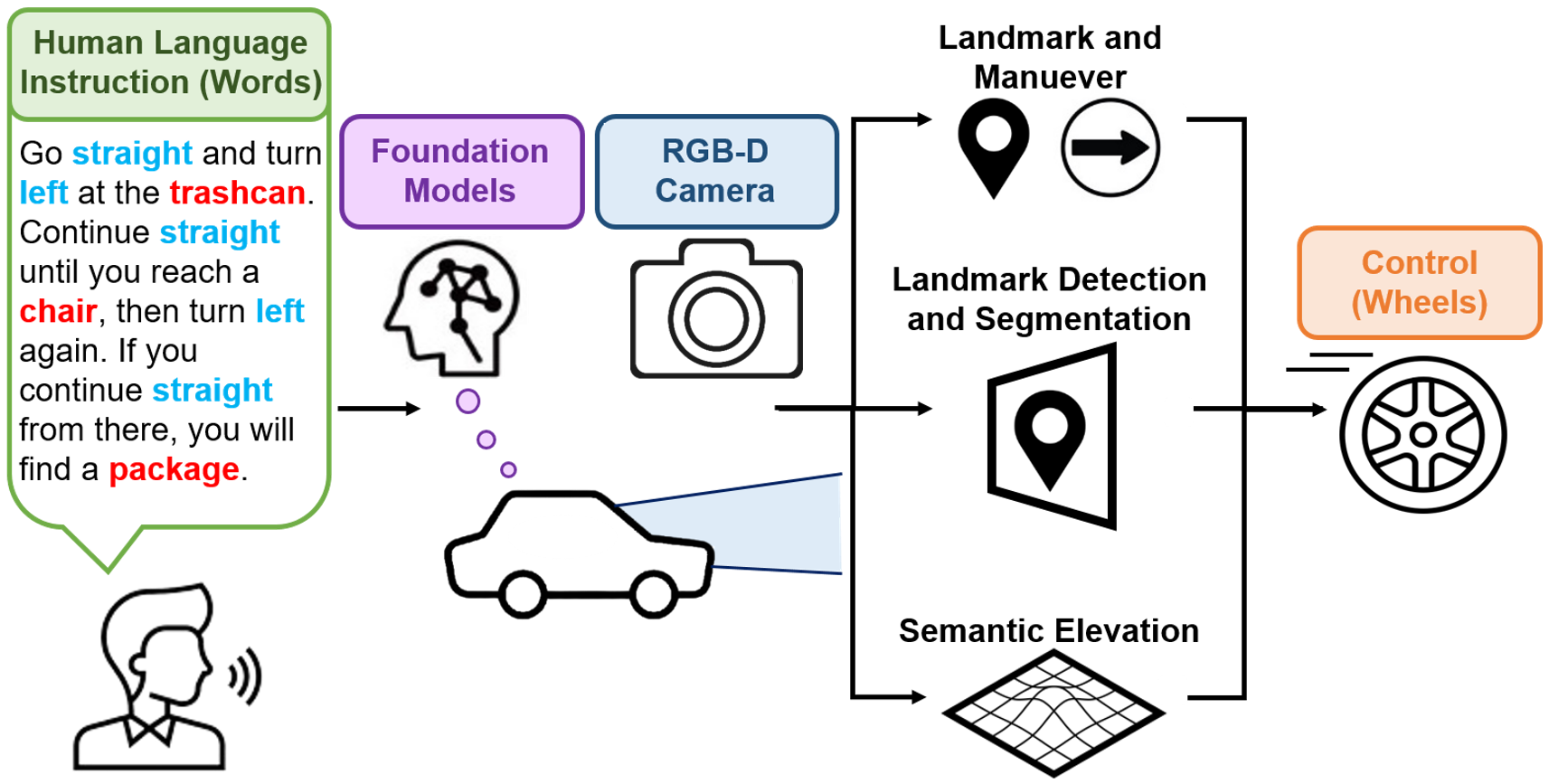}
   \caption{Overview of the Words to Wheels.}
   \label{fig:words_to_wheels}
\end{figure}
In response, this paper introduces `Words to Wheels', a novel approach that translates human language instructions into high-level maneuvers for vision-based autonomous navigation using foundation models. The process begins with an LLM transforming human instructions into a `cognitive route description', a tailored descriptive language instruction for UGVs, which is then used to extract landmarks and maneuvers. The UGV navigates using a vision-language model (VLM) while local path planning is managed through a Model Predictive Path Integral (MPPI) planner. This approach is enhanced by incorporating semantic elevation cost map, which combine elevation estimation with navigability level segmentation.

Our experiments demonstrate that our application enables an UGV to reach the destination following human language instructions in novel environments, mimicking human navigation capabilities. All computations performed onboard, except for language processing, which is handled via OpenAI’s API. Our core contributions are as follows:
\begin{figure*}[!t]
\begin{center}
    \centerline{\includegraphics[width=0.95\textwidth]{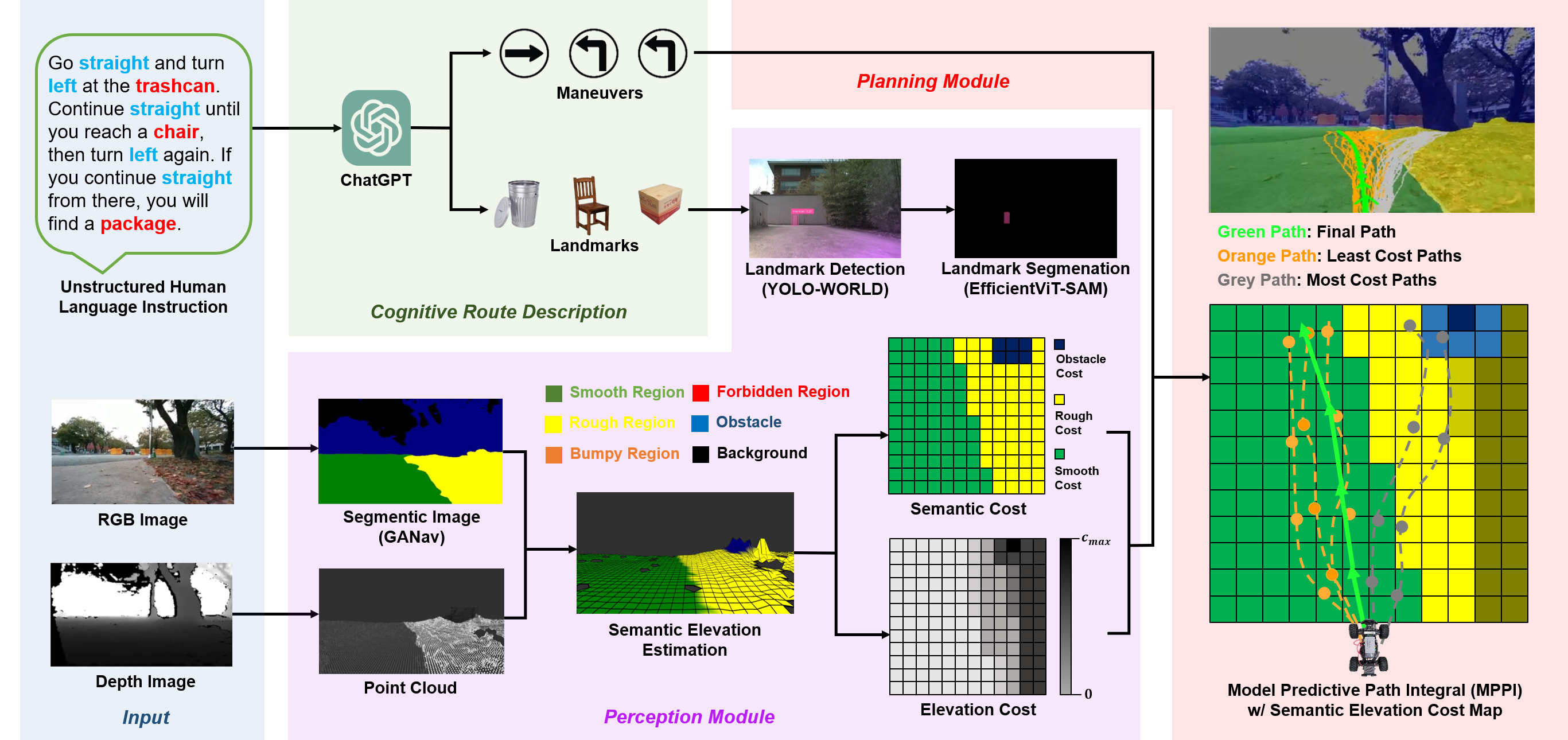}}
\end{center}
    \caption[Software Design]{The overall pipeline for vision-based autonomous driving, which interprets human language instructions using foundation models, encompasses several crucial steps. Initially, unstructured human language instructions are translated into a cognitive route description, which is then parsed into a set of maneuvers and landmarks. Depth images are utilized to generate point clouds, while RGB images are processed for navigability level segmentation using GANav. These elements contribute to creating semantic images, which, combined with point clouds, help to produce a semantic elevation cost map accounting for both semantic and elevation costs. This cost map is processed by a Model Predictive Path Integral (MPPI) planner for local planning, resulting in the output of control commands. The maneuvers are refined using the semantic elevation cost map to produce the desired actions. As the vehicle progresses, landmarks are detected and segmented using YOLO-World and EfficientViT-SAM, enabling the vehicle to autonomously verify its arrival at each landmark, seek the next one, and ultimately reach the final destination.}
    \label{fig:Overall Pipeline}
\end{figure*}
\begin{itemize}
\item We designed a novel method to effectively extract navigation commands for unmanned ground vehicles from unstructured human language by converting the instructions into a `cognitive route description'
\item We configured a visual navigation framework capable of effectively navigating novel scenes by utilizing a perception module based on zero-shot deployable foundation models and a planner module that can handle navigation maneuvers, which is compatible with the perception module.
\item We demonstrated the ability to navigate various real-world driving scenarios using solely RGB-D sensory input and onboard computing, seamlessly integrating language-based navigation instructions with a visual navigation framework, allowing it to interpret human instructions and drive autonomously without relying on GPS, pre-built maps, or prior knowledge.
\end{itemize}
\section{RELATED WORKS}
\subsection{Vision-based Autonomous Driving}
Previous research has extensively investigated vision-based autonomous driving in UGVs. One notable study \cite{rasib2021pixel} utilized pixel-level segmentation to identify areas suitable for driving on unstructured roads. This approach involved calculating the steering angle required to navigate these drivable areas using a method known as lane intersection. Furthermore, extensive research has been conducted on end-to-end visual navigation systems \cite{bojarski2016end, kahn2021badgr, shah2021ving}. These systems learn to determine the correct steering angles from images, navigate around obstacles that could cause bumps or collisions, and develop a navigational strategy based on a topological graph that combines direct visual input with planned paths. However, they still require data gathered from real-world environments and need to be trained accordingly.
\subsection{Foundation Models}
Large language models (LLMs), such as GPT \cite{dale2021gpt,sanderson2023gpt}, Gemini \cite{team2023gemini}, and LLaMa \cite{touvron2023llama}, are capable of understanding natural language and solving complex tasks via text generation. However, these are very large models, making it difficult for users to fine-tune them with custom datasets. Therefore, many researchers are exploring ways of prompt engineering, which adjusts the prompts that guide the model's response, instead of modifying the deep structural weights of the model \cite{zhao2023survey,dong2022survey,wei2022chain, si2022prompting,rong2021extrapolating}. Types of prompt engineering include `In-context learning' \cite{dong2022survey,rong2021extrapolating}, meaning that demonstrations of the task are provided to the model as part of the prompt, and `Chain of thought' \cite{wei2022chain} which involves intermediate reasoning steps to derive the final answer. 

Vision language models (VLMs), including CLIP \cite{clip}, GLIP \cite{li2022grounded, zhang2022glipv2} and LLaVa \cite{liu2024visual,liu2024improved}, extend the capabilities of LLMs to visual content by integrating computer vision and human language processing. This allows them to perform tasks involving both human language and visual content. Recently, real-time open vocabulary object detection has shown promise in applications, being capable of identifying novel categories. YOLO-World \cite{cheng2024yolo}, although limited to object detection tasks, showcases groundbreaking capabilities in real-time object detection. 

Image segmentation models are mostly based on the Segment Anything Model (SAM) \cite{kirillov2023segment}, which partitions an image into discrete groups of pixels. Recently, there have been attempts to implement SAM’s image segmentation capabilities in real-time applications \cite{zhou2023edgesam, zhang2023faster, zhao2023fast}. Meanwhile, leveraging a vision transformer-based approach \cite{zhang2024efficientvit} appears to excel.

\subsection{Language Instructed Autonomous Driving}
Previous research has explored language-instructed autonomous driving. One early study \cite{ad1} focused on advising a self-driving vehicle to pay more attention to certain objects or regions by providing human-to-vehicle advice to adjust vehicle control, thereby enhancing safety. Another paper \cite{ttv} proposed a pipeline that breaks down natural language into high-level maneuvers and generates waypoints based on these instructions. Additional studies \cite{ad2, conditionaldriving} have directly grounded the texts in driving contexts. Furthermore, recent study \cite{shah2023lm} utilized LLM to extract landmarks from given instructions, employed CLIP \cite{clip} to ground these textual landmarks in the topological map and navigate according to the topological graph. However, most of these approaches still face challenges, such as requiring at least several tens of minutes to learn to navigate in novel environments.
\section{METHODOLOGY}
\subsection{Cognitive Route Description}
As robots will receive instructions in human language from multiple individuals, handling a variety of instruction formats and styles is crucial. Different individuals may describe routes in diverse ways, leading to variations in the formats and quality of descriptions. Psychological studies analyzing the human thought process involved in describing routes often rely on visual landmarks \cite{denis1997description, michon2001and, jeanne2001role}. One of the early foundational research studies, \cite{denis1997description}, identifies an optimal procedure for producing effective route descriptions to a specific goal. This procedure can be broken down into an iteration of three steps:
\begin{itemize}
    \item $\textbf{Start Progress}$: First action to take, which can be described in the minimal form e.g. ``Go Straight Ahead"
    \item $\textbf{Announce Landmark}$: Announce the landmark that is visible from the starting point or appears during the progress.
    \item $\textbf{(Re)Orientate Addressee}$: The speaker must orient the addressee in the appropriate direction to align them with the next landmark. e.g. ``Turn Right" 
\end{itemize}
Therefore, we defined the route description that follows this rule as a `cognitive route description' (CRD). As the vehicle receives human language instructions, they are converted into CRDs using GPT-4 through three iterative steps to ensure clarity. To facilitate prompt engineering, GPT-4 is provided with multiple CRDs along with the unstructured human language instruction inputs, allowing it to accurately mimic the format—a process known as in-context learning \cite{dong2022survey,rong2021extrapolating}. Therefore, even if a vehicle is given an unstructured route description (URD), considered to be of low quality, it will convert it into a CRD to elevate the quality.

As the vehicle receives the CRD, it breaks down the description into different lists—Progress, Landmarks, and Orientations—using a prompt engineering approach that is introduced in \cite{shah2023lm}. For example, `You may go $\textbf{straight}$ to a $\textbf{trashcan}$, then turn $\textbf{left}$. After that, go $\textbf{straight}$ to a $\textbf{chair}$, then turn $\textbf{left}$ again. Then you can go $\textbf{straight}$, there is a $\textbf{box}$. Upon arrival, you may $\textbf{stop}$.' The `Extracted Progress,' `Extracted Landmarks,' and `Extracted Orientations' can be extracted as follows:
\begin{itemize}
\item Extracted Progress: {Straight, Straight, Straight}
\item Extracted Landmarks: {Trashcan, Chair, Box}
\item Extracted Orientations: {Left, Left, Stop}
\end{itemize}
Since `Extracted Progress' and `Extracted Orientations' represent actions to be taken, they are merged into a sequence of maneuvers:
\begin{itemize}
\item Extracted Maneuvers: {Straight, Left, Straight, Left, Straight, Stop}
\end{itemize}
This sequence successfully breaks down the CRD into a sequence of target landmarks and navigation commands that indicate how to reach the designated destination. As the robot sequentially identifies each target landmark, it reorients itself and begins its next maneuver while continuously seeking the next landmark. This iterative approach continues until the vehicle reaches each target landmark and receives the termination command, ‘Stop.’ 
\begin{figure}[t]
  \centering
   \includegraphics[width=0.98\linewidth]{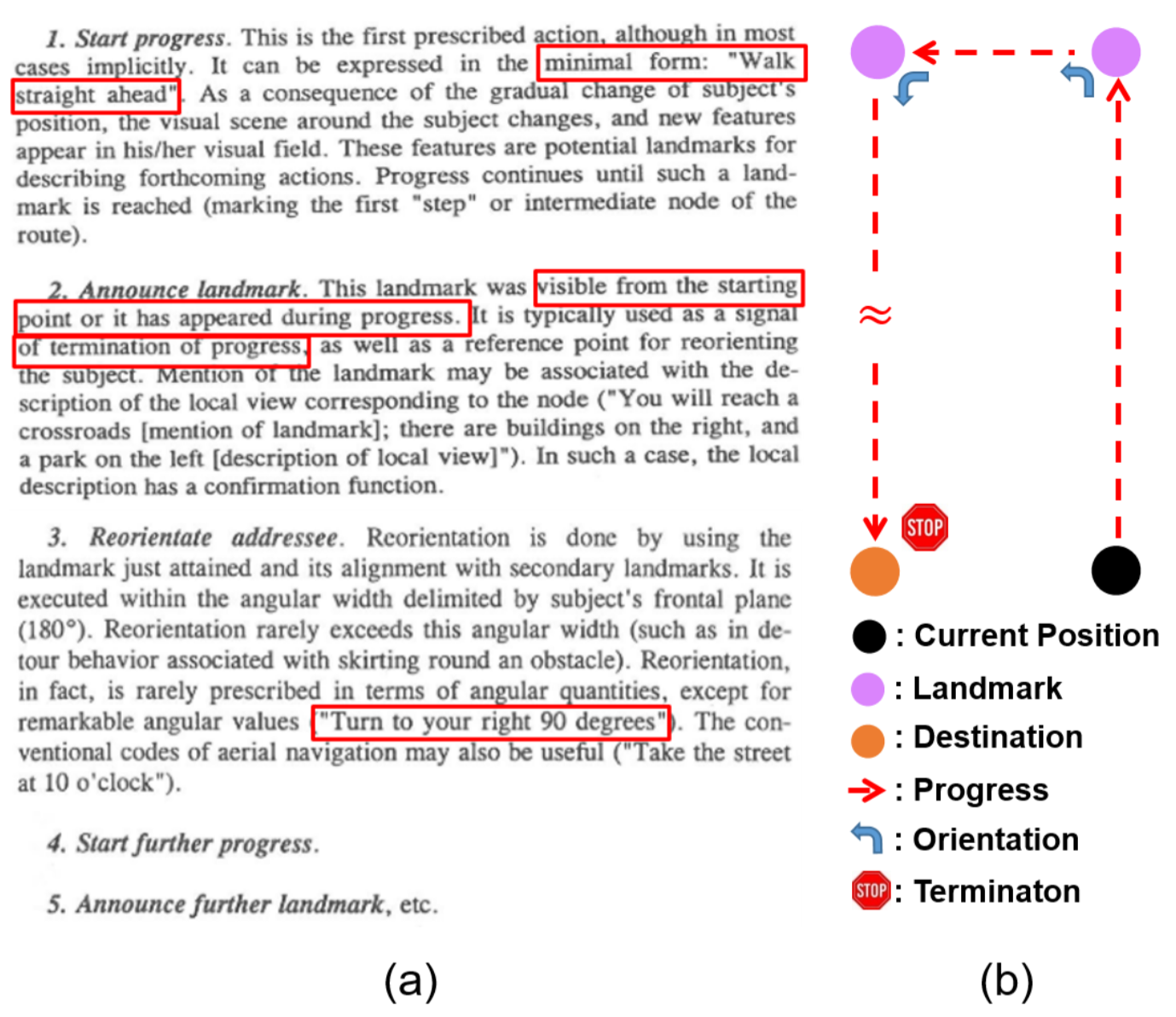}
   \caption{(a) Paragraphs extracted from `The description of routes: a cognitive approach to the production of spatial discourse.' illustrate the steps to successfully describe routes (pp. 420-421) \cite{denis1997description}. (b) Visualization of cognitive route description.}
   \label{fig:crd}
\end{figure}
\subsection{Perception Module}
\subsubsection{Landmark Detection and Segmentation}
YOLO-World\cite{cheng2024yolo} is employed for target landmark detection, capable of identifying various landmarks in real-time onboard. Upon detecting a landmark, zero-shot segmentation, EfficientViT-SAM \cite{zhang2024efficientvit}, of the detected object is applied. This method accurately segments the object within the identified bounding box, providing detailed delineation for precise distance estimation fused with the depth image. 

\subsubsection{Semantic Elevation Cost Map}
GANav \cite{9810192} is utilized to categorize navigability levels of different terrains by distinguishing between smooth, rough, bumpy, and non-navigable areas. This model is pre-trained on outdoor open datasets such as RUGD and RELLIS-3D, which feature six texture-based classes for terrain segmentation: smooth (concrete, asphalt), rough (grass, gravel), forbidden (water, bushes), obstacle and background. Although the model does not strictly guarantee domain adaptability in novel environments, the pre-trained weights demonstrate sufficient capability in such settings. We leverage this capability by assigning different navigability costs on different navigability levels, forming a semantic cost map at a specified resolution, $\mathbf{C}_{S}(x, y)$, which outputs the semantic cost at any given (x, y) coordinate.

For elevation cost estimation, robot-centric elevation estimation is employed \cite{fankhauser2014robot}, resulting in the elevation cost map, $\mathbf{C}_{E}(x,y)$. With the point cloud data and navigability information, the estimated elevation is fused with the outcomes of the navigability level segmentation. Both semantic and elevation costs are then simply aggregated to estimate the total cost, termed the semantic elevation cost map, $\mathbf{C}_{SE}(x,y)$.

\subsection{Planning Module}
\subsubsection{Maneuvers to Planning}
To execute maneuvers as instructed by the language, $\mathbf{C}_{SE}$ is dynamically modified to reflect the desired maneuver during local planning. There are four maneuver options: `Straight', `Left', `Right', and `Stop'. For `Straight' and `Stop', the cost map remains unaltered. The `Straight' maneuver allows the local planner to proceed along the existing cost map, while the `Stop' maneuver instructs the low-level controller to halt the vehicle. For the `Left' and `Right' maneuvers, $\mathbf{C}_{SE}$ is updated by imposing a penalty on the designated side of the cost map.

\begin{figure}[t]
  \centering
   \includegraphics[width=0.99\linewidth]{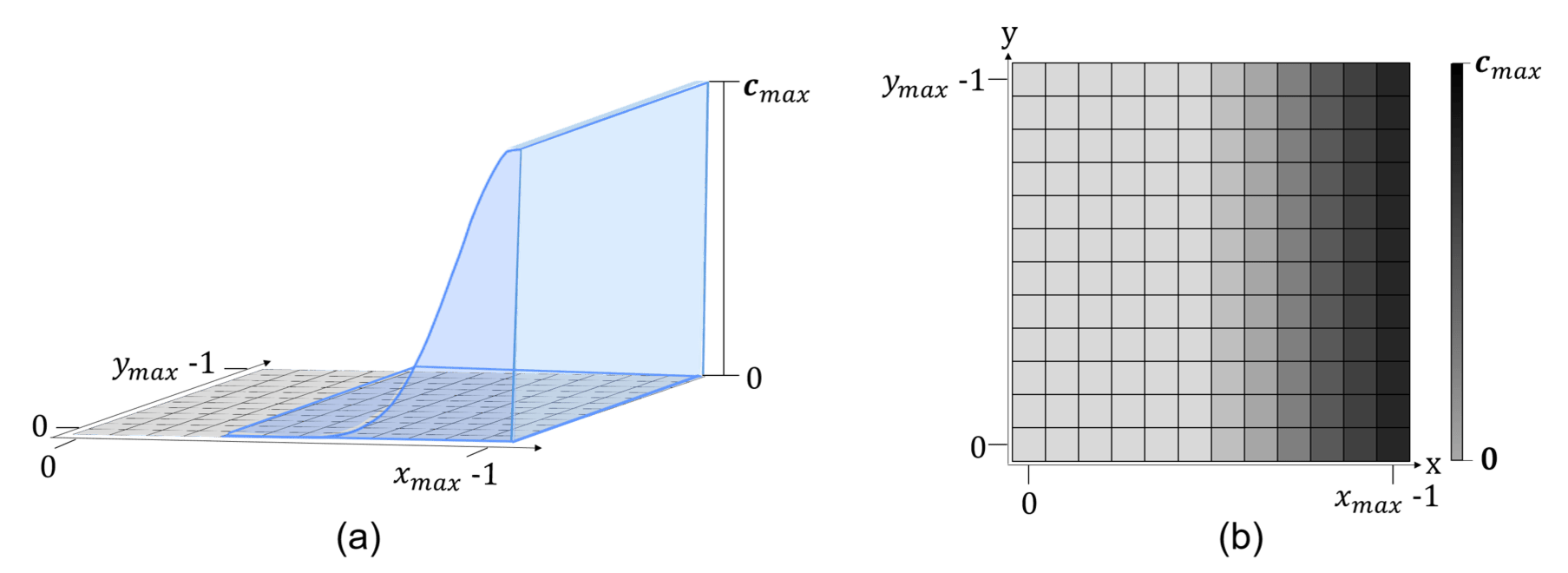}
   \caption{(a) Visualization of Gaussian distribution inspired penalty to induce left turn maneuver, (b) Updated cost map promoting left turn by penalizing right side of the cost map.}
   \label{fig:maneuver}
\end{figure}

This penalty is applied using a method inspired by a Gaussian distribution, (see Fig. \ref{fig:maneuver}), with values ranging from 0 to $\mathbf{c}_{max}$. In the case of a `Left' maneuver, the peak of the Gaussian distribution is set at the rightmost grid of the cost map, thus promoting a leftward turn, (see Alg. \ref{alg:ac}). The sharpness of the turn can be modulated by adjusting the standard deviation of the distribution, $\sigma$.

Therefore, the updated $\mathbf{C}_{SE}$ not only facilitates the intended maneuver but also integrates seamlessly with the semantic elevation cost. This ensures that the vehicle can safely execute the turn without being strictly forced to change direction, regardless of underlying semantic elevation costs.

\subsubsection{MPPI-Based Local Planning}
\begin{algorithm}
\caption{Update Cost Map for Turn Maneuvers}
\begin{algorithmic}[]
\State \textbf{Input:} 
\State $\mathbf{C}_{SE}$: Input semantic elevation cost map
\State $\mathbf{x}_{max}, \mathbf{y}_{max}$: x and y dimension of $\mathbf{C}_{SE}$ 
\State $\mathbf{c}_{max}$: Maximum cost of $\mathbf{C}_{SE}$
\State $\sigma$: Desired standard deviation
\State $\text{M}$: Input maneuver 
\If{M = `Left'}
    \State \( \mu = \mathbf{x}_{max} - 1 \)
\ElsIf{M = `Right'}
    \State \( \mu = 0\)
\Else \Comment{M = `Straight' or `Stop'}
\State\textbf{return} $\mathbf{C}_{SE}$\  
\EndIf
\For {\( x = 0 \) to \( \mathbf{x}_{max} - 1 \)}
    \For {\( y = 0 \) to \( \mathbf{y}_{max} - 1 \)}
        \State \( \mathbf{C}_{SE}(x,y) \mathrel{+}= \mathbf{c}_{max} e^{-\frac{(x - \mu)^2}{2\sigma^2}} \)
        \State \( \mathbf{C}_{SE}(x,y) = \max(\mathbf{C}_{SE}(x,y), \mathbf{c}_{max}) \)
    \EndFor
\EndFor
\State\textbf{return} $\mathbf{C}_{SE}$\
\end{algorithmic}
\label{alg:ac}
\end{algorithm}
$\mathbf{C}_{SE}$ with a resolution of 10 cm is provided to the MPPI planner \cite{williams2018information}, which generates 5,000 paths, each comprising 20 time steps, at a frequency of 40Hz. The cost for each path is calculated at each time step, referring to the $\mathbf{C}_{SE}$. Subsequently, the MPPI planner performs a weighted summation, taking into account the cost assigned to each path using the conventional MPPI weighted summation approach, and then outputs the final path. Once the target landmark is detected while traveling, the distance to the landmark can be estimated using the semantic image of the landmark and the corresponding depth image. If this distance falls below a predetermined threshold, the vehicle is considered to have reached the landmark. Subsequently, the vehicle proceeds to the next maneuver (orientation) and the updated cost inducing the maneuver persists while detecting the landmark, with a buffer duration after the landmark is no longer detected. If the vehicle has not reached the landmark, it performs visual servoing to align its heading towards the landmark. This maneuver also undergoes a collision-checking process on the $\mathbf{C}_{SE}$ for validation to prevent potential collisions.
\begin{figure}[t]
  \centering
   \includegraphics[width=0.92\linewidth]{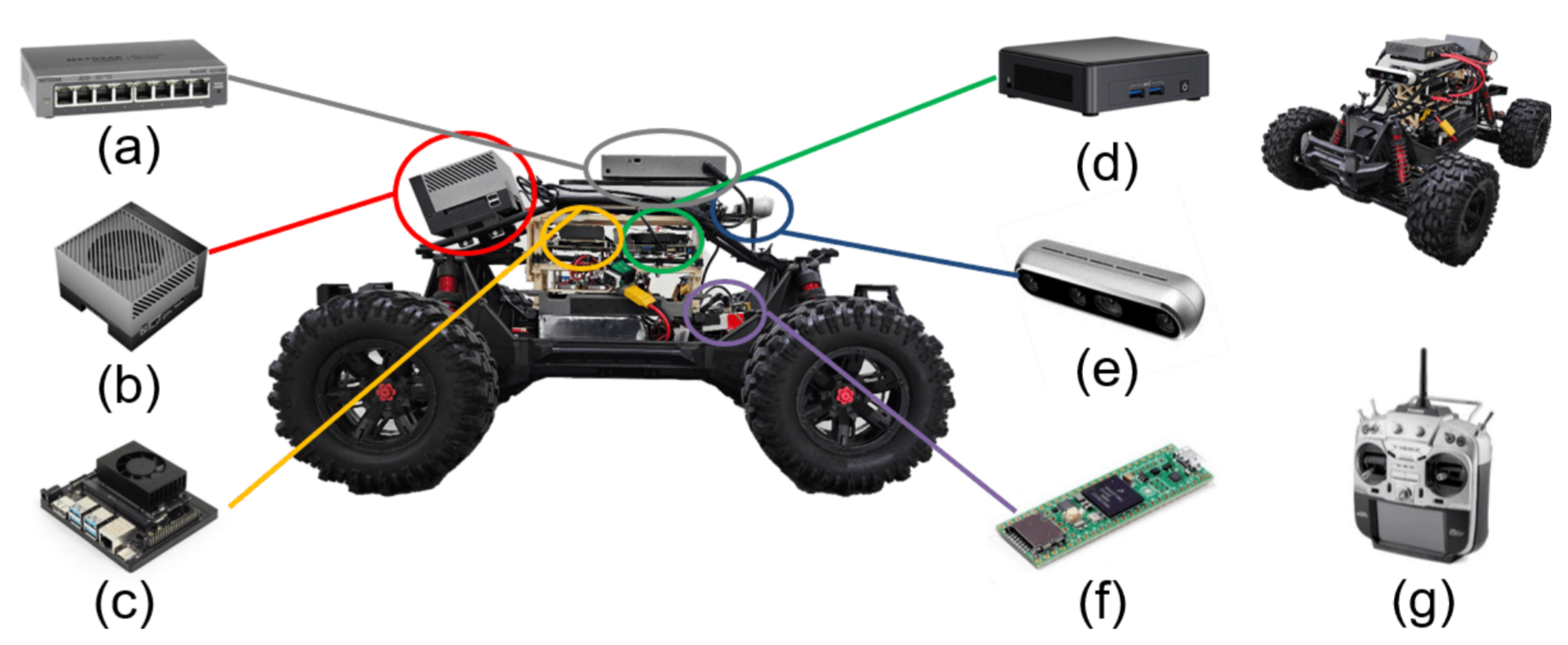}
   \caption{Hardware design of an unmanned ground vehicle (32"$\times$20"$\times$16"), Traxxas buggy car platform, includes: (a) Nvidia Jetson Orin NX, (b) Nvidia Jetson Orin AGX, (c) NETGEAR GS108E,  (d) Intel NUC 11 Pro i7, (e) Intel RealSense D455, (f) Teensey 4.1 Development Board, (g) Futaba T18SZ}
   \label{fig:Hardware}
\end{figure}
\section{EXPERIMENTS}
\subsection{Environmental Setup}
\begin{figure*}[!t]
\centering
    \includegraphics[height=0.162\textwidth,valign=c]{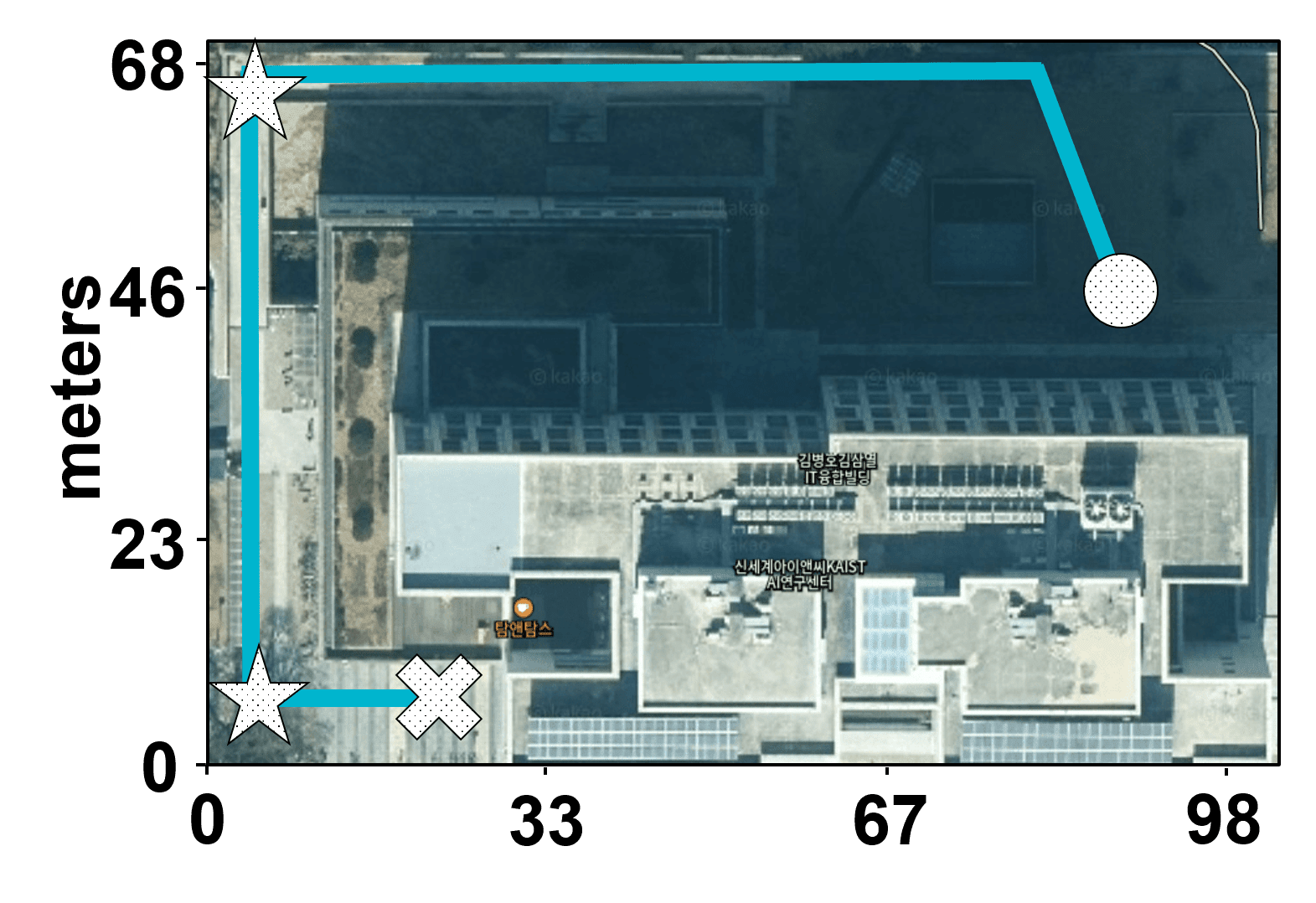}
    \includegraphics[width=0.184\textwidth,valign=c]{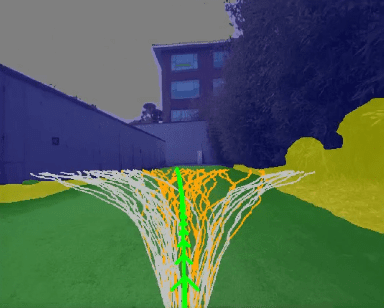}
    \includegraphics[width=0.184\textwidth,valign=c]{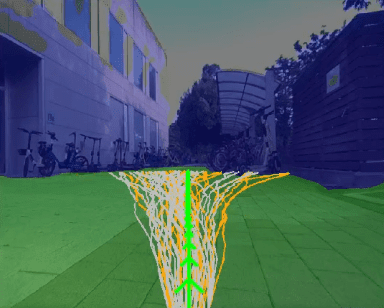}
    \includegraphics[width=0.184\textwidth,valign=c]{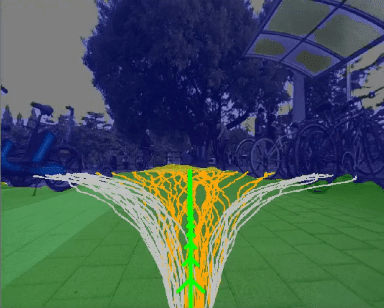}
    \includegraphics[width=0.184\textwidth,valign=c]{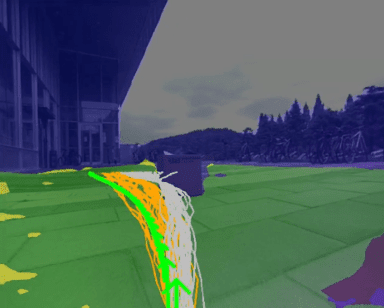}
    \includegraphics[height=0.162\textwidth,valign=c]{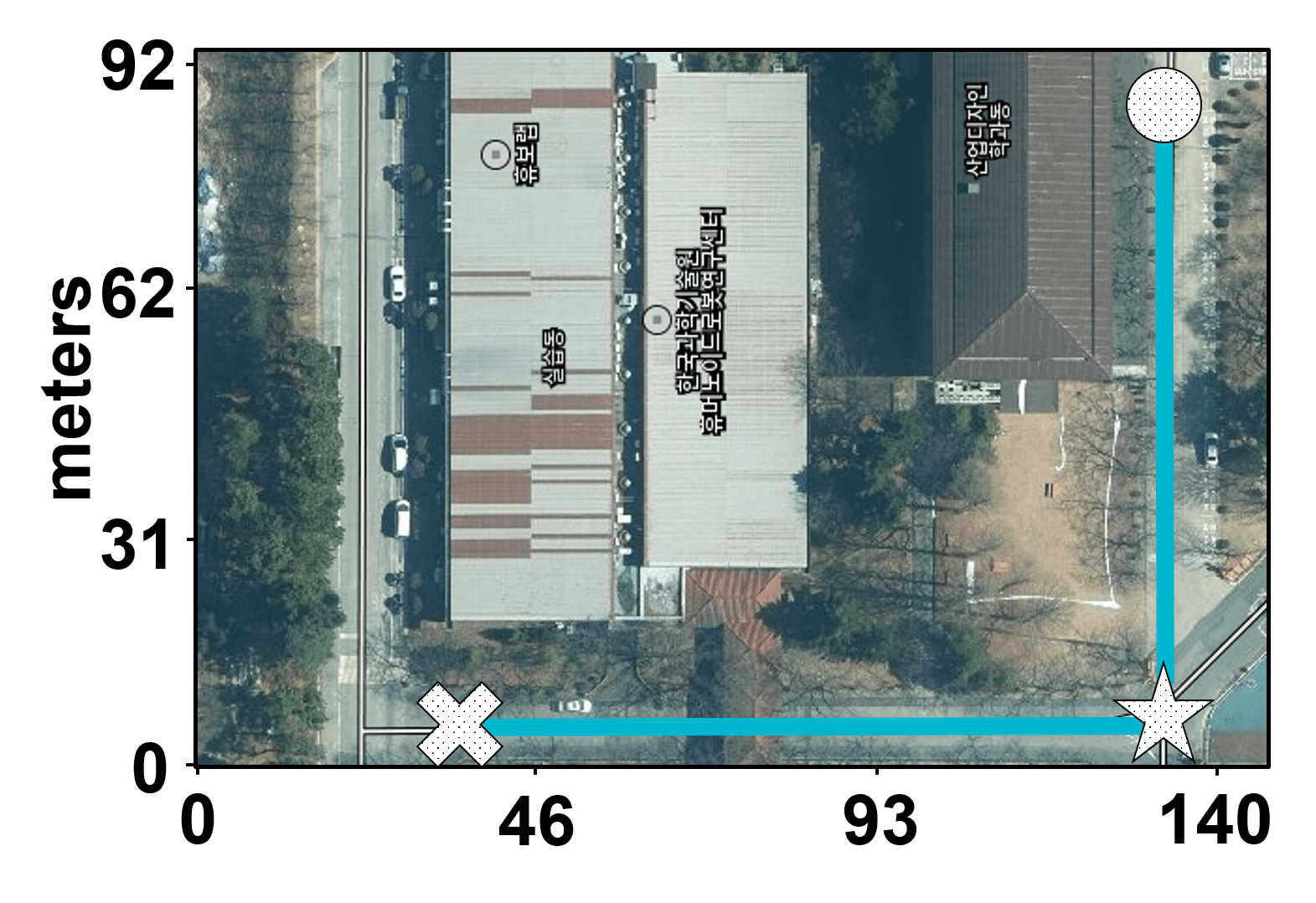}
    \includegraphics[width=0.184\textwidth,valign=c]{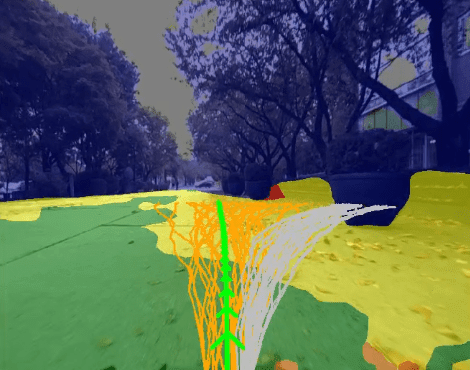}
    \includegraphics[width=0.184\textwidth,valign=c]{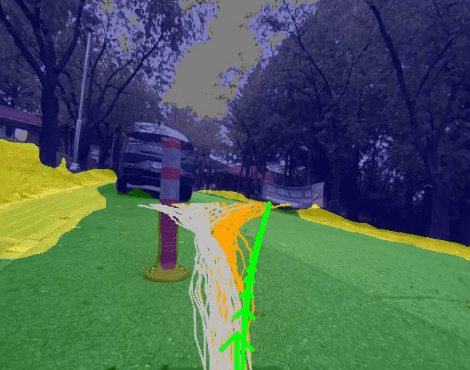}
    \includegraphics[width=0.184\textwidth,valign=c]{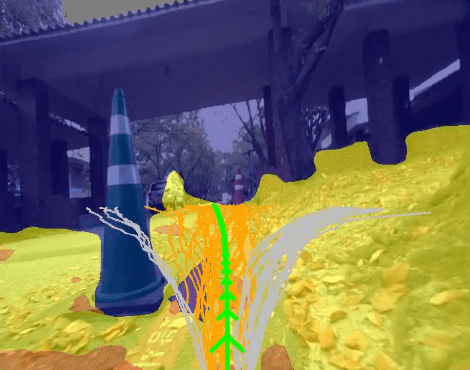}
    \includegraphics[width=0.184\textwidth,valign=c]{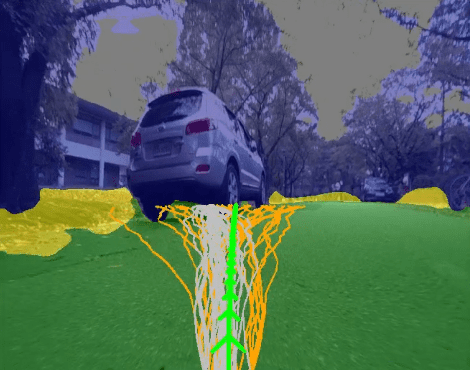}
    \includegraphics[height=0.162\textwidth,valign=c]{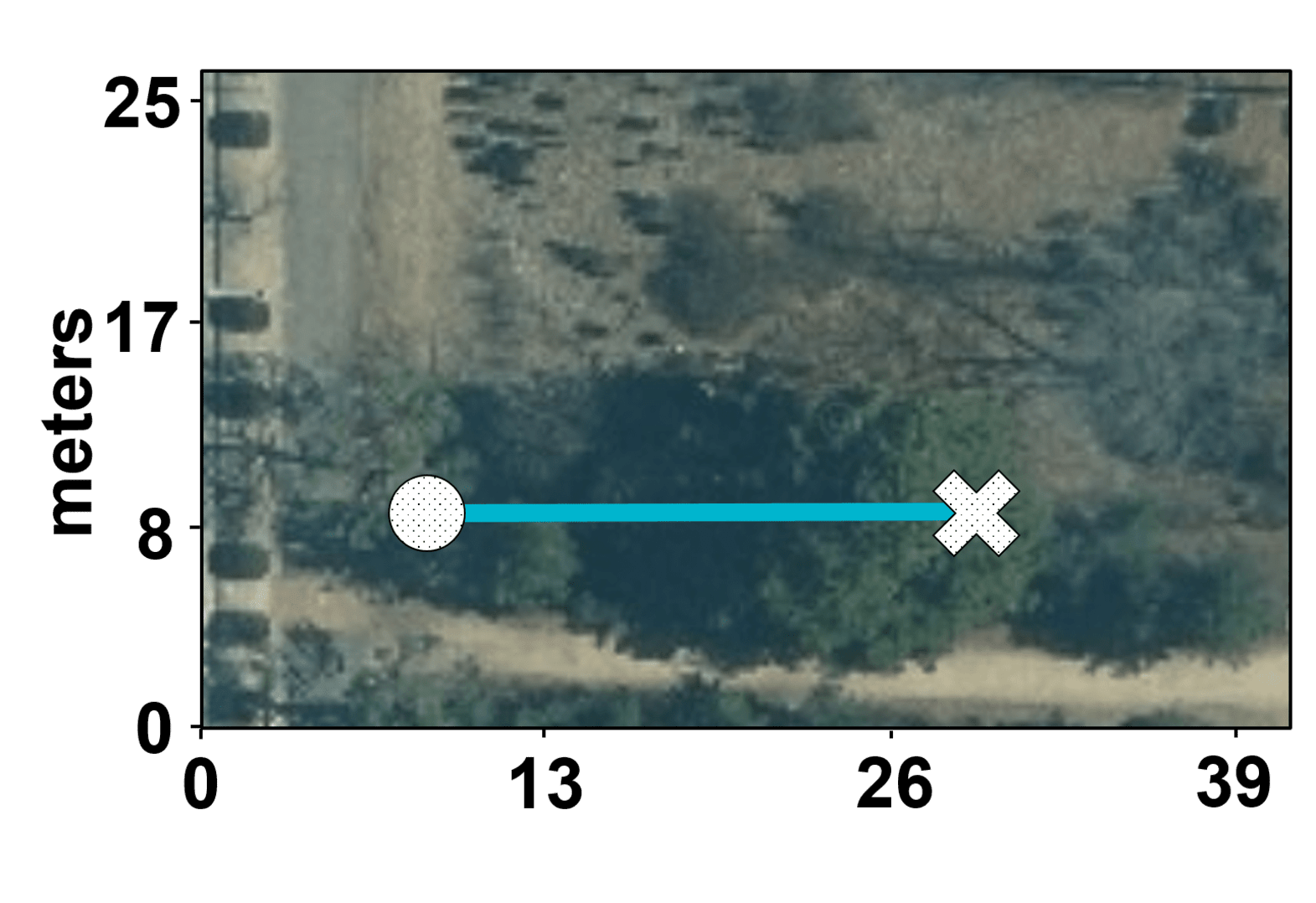}
    \includegraphics[width=0.184\textwidth,valign=c]{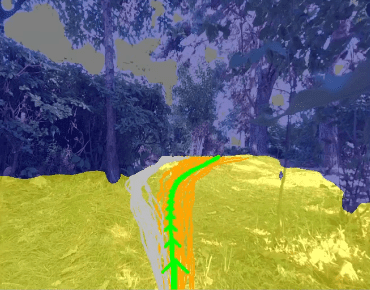}
    \includegraphics[width=0.184\textwidth,valign=c]{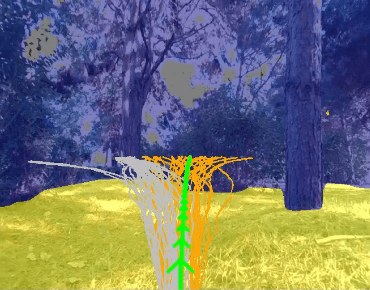}
    \includegraphics[width=0.184\textwidth,valign=c]{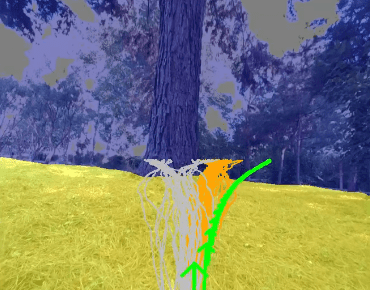}
    \includegraphics[width=0.184\textwidth,valign=c]{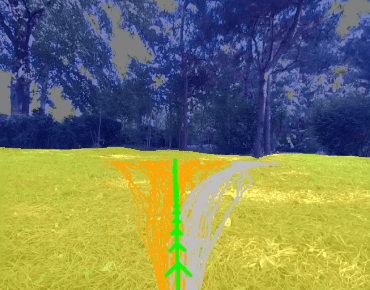}
    \caption[Scenarios]{The dots represent the starting points, the cross-shaped patterns indicate the goal points (last landmarks), the stars denote landmarks, and the edges represent routes. In the first-person view (FPV) images, different levels of drivable regions are overlaid in different colors. (Green: Smooth, Yellow: Rough, Orange: Bumpy, Red: Forbidden, Blue: Obstacle, Black: Background). Additionally, a few of the paths with the lowest costs (orange), a few with the highest costs (grey), and the final path (green arrow) derived from the weighted sum are visualized.
First row: outdoor delivery service scenario, approximately 180 meters.
Second row: Another outdoor delivery service scenario, approximately 190 meters.
Third row: off-road observing scenario, approximately 20 meters.}
    \label{fig:Scenarios}
\end{figure*}
\subsubsection{Hardware Configuration}
As shown in Fig. \ref{fig:Hardware}, the overall system configuration utilizes a Traxxas buggy car platform equipped with three onboard embedded computing devices: an Intel NUC11 Pro i7, an Nvidia Jetson AGX Orin, and an Nvidia Jetson Orin NX. Communication among these components is facilitated through a NETGEAR GS108E switch via Ethernet. For RGB and depth image acquisition, we employed an Intel RealSense D455 camera. Additionally, a Teensy 4.1 development board has been implemented to manage the low-level controllers.
\subsubsection{Model Configuration}
We utilized multiple pre-trained models, including LLM, object detection, object segmentation, and navigability level segmentation. GPT-4 was initially executed to process and understand URD through OpenAI's API, while tasks such as object detection, object segmentation, and navigability level segmentation were processed on onboard computing devices. The frequencies at which these processes were run are listed in Table \ref{tab:model}. Object detection and segmentation are integrated into a single pipeline and therefore operate at the same frequency.
\begin{table}[t]
\begin{center}
\caption{Models employed in Words to Wheels}
\footnotesize
    \begin{tabular}{l|l|l} 
    \hline 
    Purpose & Model & Frequency \\
    \hline \hline
    Route Understanding & GPT-4 & N/A  \\ 
    \hline
    \raisebox{-1.25ex}[0pt][0pt]{Object Detection} & YOLO-World-L, & \\ 
    & CLIP-ViT-B-32 & 10Hz\\ 
    \cline{1-2}
    Object Segmentation & EfficientViT-SAM-L0 &  \\ 
    \hline
    Navigability Level Segmentation & GANav-RUGD-6 & 20Hz \\
    \hline
\end{tabular}
\label{tab:model}
\end{center}
\end{table}

\subsubsection{Scenario Configuration}
\paragraph{Scenario 1}
The outdoor experiment simulates a postal service, starting at the building's back and ending at the front where a package is located. The language instruction is: `You may go $\textbf{straight}$ to a $\textbf{trashcan}$, then turn $\textbf{left}$. After that, go $\textbf{straight}$ to a $\textbf{chair}$, then turn $\textbf{left}$ again. Then, there is a $\textbf{package}$. Upon arrival, you may $\textbf{stop}$.’
\begin{itemize}
\item Desired Landmarks: {Trashcan, Chair, Box}
\item Desired Maneuvers: {Straight, Left, Straight, Left, Straight, Stop}
\end{itemize}
\paragraph{Scenario 2}
Another experiment simulates a postal service over a larger area, starting in a parking lot and ending near a package in a driveway by another building. The language instruction is: `Go $\textbf{straight}$ to an $\textbf{orange cone}$, then turn $\textbf{right}$. After that, go $\textbf{straight}$ to a $\textbf{package}$.'
\begin{itemize}
\item Desired Landmarks: {Orange Cone, Package}
\item Desired Maneuvers: {Straight, Right, Straight, Stop}
\end{itemize}
\paragraph{Scenario 3}
The off-road experiment is designed to simulate an observing scenario. The language instruction is: `Find a $\textbf{white\ ball}$, and $\textbf{stop}$.'
\begin{itemize}
\item Desired Landmark: {White Ball}
\item Desired Maneuvers: {Straight, Stop}
\end{itemize}
\section{Main Results}
\subsection{Route Understanding Performance with Comparison Study} 
This experiment aims to visualize the effectiveness of converting URD into CRD and to compare the performance of different models. The URDs, consisting of landmarks and maneuvers, were composed by ChatGPT under the assumption that the starting position is aligned with the next landmark.

An ideal route description should include N landmarks, N progress, and N orientations, with the final orientation being a `Stop' to indicate route termination, as illustrated in Fig. \ref{fig:crd}. The average score is calculated by dividing the total number of landmarks, progress indicators, and orientations (3N) by their respective sums. The average progress quantifies the proportion of the journey completed before encountering a missing component in the route description. It is computed by dividing the number of landmarks reached prior to the missing component by the total number of landmarks.

Initially, we assessed the likelihood of the URD omitting meaningful information. As shown in the table, it is indeed likely to miss important details. Next, we applied various language models to convert the URD into CRD, using the same prompt engineering and in-context learning methodology as previously described. While all language models demonstrated acceptable performance, GPT-4 outperformed the others in these metrics, Table \ref{tab:Language Model Comparison}. Therefore, GPT-4 was employed for the real-world experiments.
\begin{table}[t]
\begin{center}
    \caption[Route Understanding Performance with Comparison Study]{Route Understanding Performance with Comparison Study}
    \begin{tabular}{l|l|l}
    \hline
    Method & Avg. Score & Avg. Progress \\ \hline \hline
    URD & 0.76 & 0.53 \\ \hline
    CRD (GPT-3.5-turbo) & 0.91 & 0.73 \\ \hline
    CRD (GPT-4o) & 0.93 & 0.81 \\ \hline
    \bf{CRD (GPT-4)} & \bf{0.98} & \bf{0.90} \\ \hline
    \end{tabular}
    \label{tab:Language Model Comparison}
    \end{center}
\end{table}
\subsection{Autonomous Driving Performance with Ablation Study}
We initially tested our application's performance in earlier scenarios (see Fig. \ref{fig:Scenarios}). Additionally, we ablated key components in our visual navigation framework, including the $\mathbf{C}_{SE}$ and the MPPI, to assess their effectiveness. Replacing the $\mathbf{C}_{SE}$ with a height-based obstacle avoidance approach led to a significant decline in both the success rate and average progress, as shown in Table \ref{tab:t}. The height-based approach solely focuses on geometric hazards and fails to differentiate between navigable and non-navigable areas based on semantic classes. This oversight can result in accidental incursions into undesired areas—a critical concern when the terrain includes non-geometric hazards such as water. Moreover, this method does not distinguish between surfaces like grass, gravel, concrete, or asphalt, increasing the likelihood of deviation from the path intended by the human instructor.

Furthermore, when we replaced the MPPI planner module with a motion primitives planner \cite{garrote2014rrt}, which selects the least-cost path from pre-generated paths, we observed a similar decline in both success rate and average progress. The motion primitives approach is highly sensitive to noise. Although navigability level segmentation can provide an adequate ability to distinguish between these classes, it may still yield noisy results, especially in unfamiliar environments. In scenarios with noisy segmentation, the motion primitives might depend on inaccurately segmented regions, potentially leading to the selection of a path that results in collisions.

Our method integrates both the $\mathbf{C}_{SE}$ and the MPPI, leveraging the strengths of each component to achieve the highest success rate and average progress. This approach allows the system to effectively navigate complex environments by considering terrain textures and incorporating noise as a factor, while maintaining an emphasis on general trends. In most failure cases, the UGV fails to detect or falsely detect the desired landmark.
\begin{table}[t]
\begin{center}
    \caption[Autonomous Driving Performance with Ablation Study]{Autonomous Driving Performance with Ablation Study}
    \begin{tabular}{l|l|l}
    \hline
    Method & Success Rate & Avg. Progress \\ \hline \hline
    Ours (-$\mathbf{C}_{SE}$) & 0.2 & 0.51 \\ \hline
    Ours (-MPPI)& 0.3 & 0.56 \\ \hline
    \bf{Words To Wheels (Ours)} & \bf{0.7} & \bf{0.83} \\ \hline
    \end{tabular}
    \label{tab:t}
    \end{center}
\end{table}
\begin{figure}[t]
  \centering
   \includegraphics[width=0.99\linewidth]{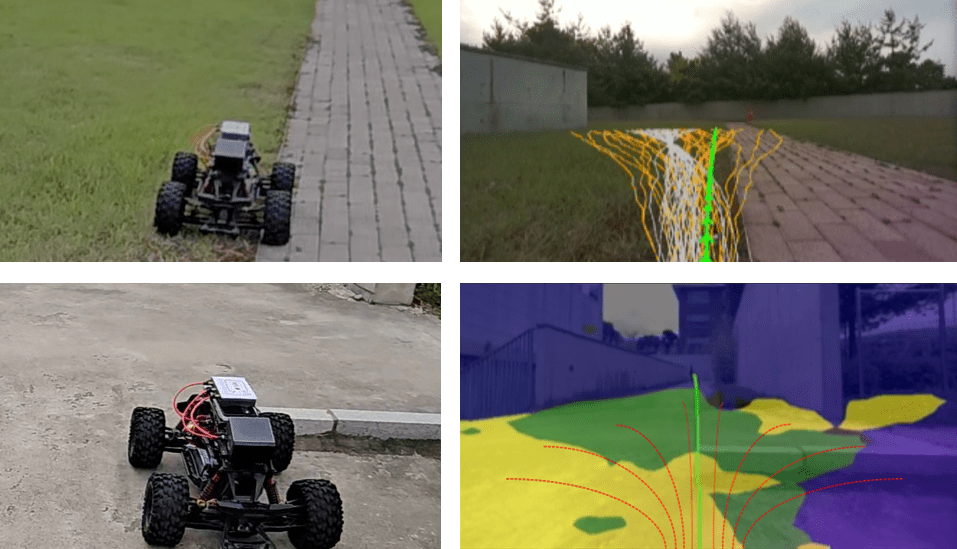}
   \caption{First row: Cost map ablation study, the height-based cost map treats concrete and grass the same, diverging from the desired domain. Second row: Planning ablation study, in the case of noisy segmentation, the motion primitives are prone to noise, which could lead to potential collisions. (Green path: selected path; Red path: motion primitives).}
   \label{fig:ablation}
\end{figure}
\section{CONCLUSIONS}
In this study, we presented `Words to Wheels,' an innovative application of foundation models that empowers UGVs to navigate towards specified destinations based on verbal instructions in unfamiliar settings. Our approach involved translating unstructured human language into a `cognitive route description,' which bridges the gap between natural language instruction and robotic navigation. Additionally, we established a visual navigation framework tailored for traversing complex environments with the given instructions, leveraging zero-shot deployable foundation models with onboard computing capabilities. Our experiments demonstrated the system's ability to navigate diverse real-world driving scenarios using only RGB-D images as sensory inputs. By incorporating language-based instructions into the visual navigation framework, our system could interpret and execute these commands effectively, thereby facilitating autonomous driving to the designated destination without depending on GPS or pre-acquired information in various novel environments. Although our approach necessitates a route description based on visual landmarks, these developments are expected to enhance the versatility and broaden the application potential of UGVs, creating more adaptable and reliable options for a variety of real-world applications.

\section*{ACKNOWLEDGMENT}
We thank Hyunwoo Nam, Juhyeong Roh, and Jehun Kang from the Korea Advanced Institute of Science and Technology (KAIST) for their assistance with the real-world experiments.

\newpage
\bibliographystyle{unsrt}
\bibliography{IEEEexample.bib}
\end{document}